\documentclass[sigconf]{acmart}
\usepackage{graphicx}
\usepackage{amsmath}
\usepackage{amsthm}
\usepackage{amsfonts}
\usepackage{bm}
\usepackage{caption}
\usepackage{multirow}
\usepackage[english]{babel}
\usepackage{adjustbox}
\usepackage{subfigure}
\usepackage{booktabs}
\usepackage{array}
\usepackage{algorithm}

\usepackage{algpseudocode}
\usepackage{multicol}
\usepackage{tabu}
\usepackage{caption}
\floatname{algorithm}{Algorithm}

\AtBeginDocument{%
  \providecommand\BibTeX{{%
    \normalfont B\kern-0.5em{\scshape i\kern-0.25em b}\kern-0.8em\TeX}}}
\copyrightyear{2019} 
\acmYear{2019} 
\setcopyright{acmcopyright}
\acmConference[SIGIR '19]{Proceedings of the 42nd International ACM SIGIR Conference on Research and Development in Information Retrieval}{July 21--25, 2019}{Paris, France}
\acmBooktitle{Proceedings of the 42nd International ACM SIGIR Conference on Research and Development in Information Retrieval (SIGIR '19), July 21--25, 2019, Paris, France}
\acmPrice{15.00}
\acmDOI{10.1145/3331184.3331283}
\acmISBN{978-1-4503-6172-9/19/07}

\begin{document}
\fancyhead{}
\title{Graph Intention Network for Click-through Rate Prediction in Sponsored Search}

  \author{
	Feng Li,  Zhenrui Chen, Pengjie Wang*,	Yi Ren, Di Zhang, Xiaoyu Zhu
 }
  \affiliation{%
   \institution{Alibaba Group}
 }
 \email{
 {adam.lf, zhenrui.czr, pengjie.wpj, hengrui.ry, di.zhangd, benjamin.zxy}@alibaba-inc.com
 }
\thanks{*This author is the one who gives a lot of guidance in the work.}

\begin{abstract}
Estimating click-through rate (CTR) accurately has an essential impact on improving user experience and revenue in sponsored search. For CTR prediction model, it is necessary to make out user’s real-time search intention. Most of the current work is to mine their intentions based on users' real-time behaviors. However, it is difficult to capture the intention when user behaviors are sparse, causing the \textit{\textbf{behavior sparsity}} problem. Moreover, it is difficult for user to jump out of their specific historical behaviors for possible interest exploration, namely \textit{\textbf{weak generalization}} problem. We propose a new approach Graph Intention Network (GIN) based on co-occurrence commodity graph to mine user intention. By adopting multi-layered graph diffusion, GIN enriches user behaviors to solve the behavior sparsity problem. By introducing co-occurrence relationship of commodities to explore the potential preferences, the weak generalization problem is also alleviated. To the best of our knowledge, the GIN method is the first to introduce graph learning for user intention mining in CTR prediction and propose end-to-end joint training of graph learning and CTR prediction tasks in sponsored search. At present, GIN has achieved excellent offline results on the real-world data of the e-commerce platform outperforming existing deep learning models, and has been running stable tests online and achieved significant CTR improvements.
\end{abstract}


\keywords{
sponsored search, click-through rate prediction, graph neural network, intention mining
}
\maketitle

\section{Introduction}
In sponsored search, estimating click-through rate accurately is essential to improve revenue and user experience. For accurate estimation of CTR, It is critical to understand user's real-time search intentions in the CTR prediction task, because the majority of users do not describe their search intention completely through query.

Currently, lots of user intention mining method is proposed. Temporal Deep Structured Semantic Model (TDSSM) \cite{song2016multi} characterizes user's intention as long-term and short-term to capture their preference and real-time intention. Dynamic REcurrent bAsket Model (DREAM) \cite{yu2016dynamic} uses recurrent neural network (RNN) to model user's behavior sequence to improve user intention expression. Furthermore, Deep Interest Network (DIN) \cite{zhou2018deep} indicates that user interest is diverse, and uses the attention mechanism to calculate the relevance between the current advertising commodity and historical commodities clicked by the user.

However, these intention recognition methods mentioned above mainly focus on user's historical behaviors, i.e., user's intention is summarized according to historical behaviors. This kind of methods have two disadvantages: \textit{\textbf{behavior sparsity}} and \textit{\textbf{weak generalization}}. Behavior sparsity means that it is difficult to capture the user's real-time intention when user's behavior is sparse. Weak generalization refers to the user's inability to jump out of their specific historical behavior for possible interest exploration.

In addition, some graph embedding methods are introduced into the CTR prediction task by a two-stage approach. \cite{wang2018billion} uses DeepWalk \cite{perozzi2014deepwalk} to generate node sequence and the Skip-Gram model is used for graph embedding. Then, the learned node representation is further used in the CTR predict task. There are numerous work proposed for graph embedding. Graph Convolutional Network (GCN) \cite{defferrard2016convolutional} aggregates neighbor nodes through mean-pooling and generates new representations with the current nodes through nonlinear functions. Graph Attention Network (GAT) \cite{velivckovic2017graph} further proposed attention-based neighbor aggregation by calculating the correlation between the current node and neighbors.

These graph embedding based methods have achieved significant results, but these methods are not directly optimized for specific CTR prediction task, which means that these methods firstly learn graph node representation by unsupervised or semi-supervised methods and then use the learned node representations to predict the CTR. This kind of training methods is not optimized for the final goals, and node representations are not adjusted by the specific tasks, thus becoming the bottleneck of the expression ability in the CTR prediction task.

We propose a new approach Graph Intention Network (GIN) based on co-occurrence commodity graph to solve these problems. Firstly, the GIN method enriches user's behavior by multi-layered graph diffusion of user historical behaviors, and solves the behavior sparsity problem. Secondly, the weak generalization problem is alleviated by introducing co-occurrence relationship of commodities to explore the potential preferences of users. Finally, we combine this intention mining method based on co-occurrence commodity graph with the CTR prediction task by end-to-end joint training.

The main contributions of this paper are as follows.

\begin{itemize}
\item The end-to-end joint training of graph learning and CTR prediction tasks is proposed for the first time in the sponsored search ranking model.
\item The behavior sparsity and weak generalization problems are alleviated by the multi-layered intention diffusion and aggregation based on the co-occurrence click relationship graph.
\item The effectiveness of the proposed GIN method is verified by offline and online experiments.
\end{itemize}

\section{The proposed approach}

\begin{figure}[!htbp]
\centering
\includegraphics[width=0.48\textwidth, keepaspectratio]{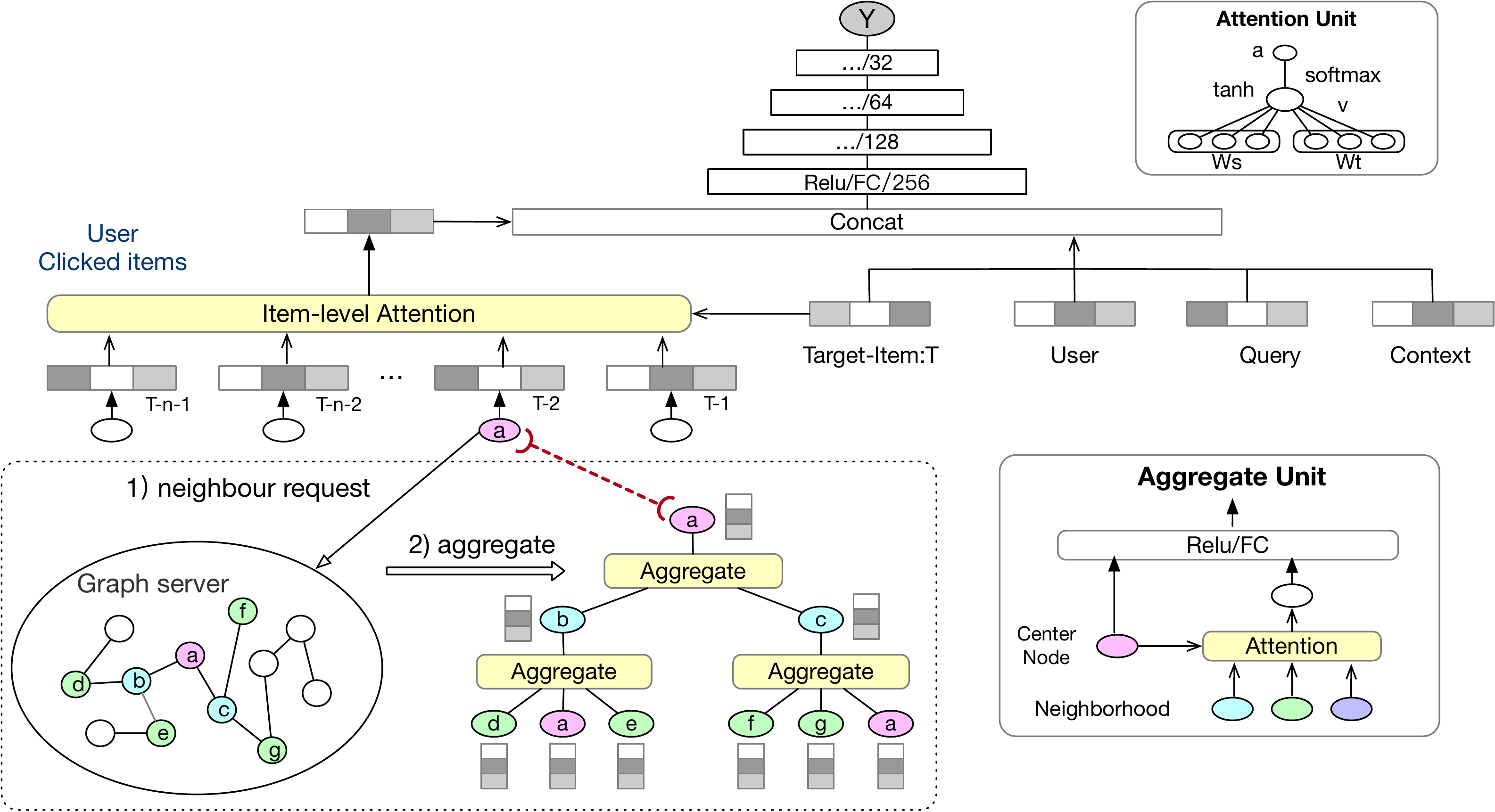}
\caption{The proposed end-to-end joint training method combines graph-based intention mining with CTR prediction tasks.
Each historical clicked sample first performs a multi-layered neighbor query on the graph service, and the attention mechanism is used to perform neighbor aggregation according to correlations between the current node and the neighbor nodes. Finally, the aggregated intention results and other features are concatenated as inputs for CTR prediction.}
\label{fig:fig1}
\end{figure}

In this section, we introduce the GIN method in detail, as shown in Fig. \ref{fig:fig1}. Firstly, the construction of co-occurrence commodity graph based on historical behaviors is introduced. Secondly, how to diffuse and aggregate multiple layers of implicit intention is introduced based on the co-occurrence commodity graph. Finally, the end-to-end joint training method is presented to combine the graph-based intention mining with CTR prediction tasks.

\subsection{Graph construction}

User historical clicks are regarded as a sequence, only click behaviors in the last month was intercepted to balance performance and effectiveness. The behavior sequences are segmented into sessions based on query similarity to prevent edge construction across dissimilar queries. Each commodity in the session constructs several undirected edges by window size, thus constructing a co-occurrence commodity graph. The node type is commodity only, and the weight of edge indicates the number of co-occurrence times.

\begin{figure}[!htbp]
\centering
\includegraphics[width=0.40\textwidth, keepaspectratio]{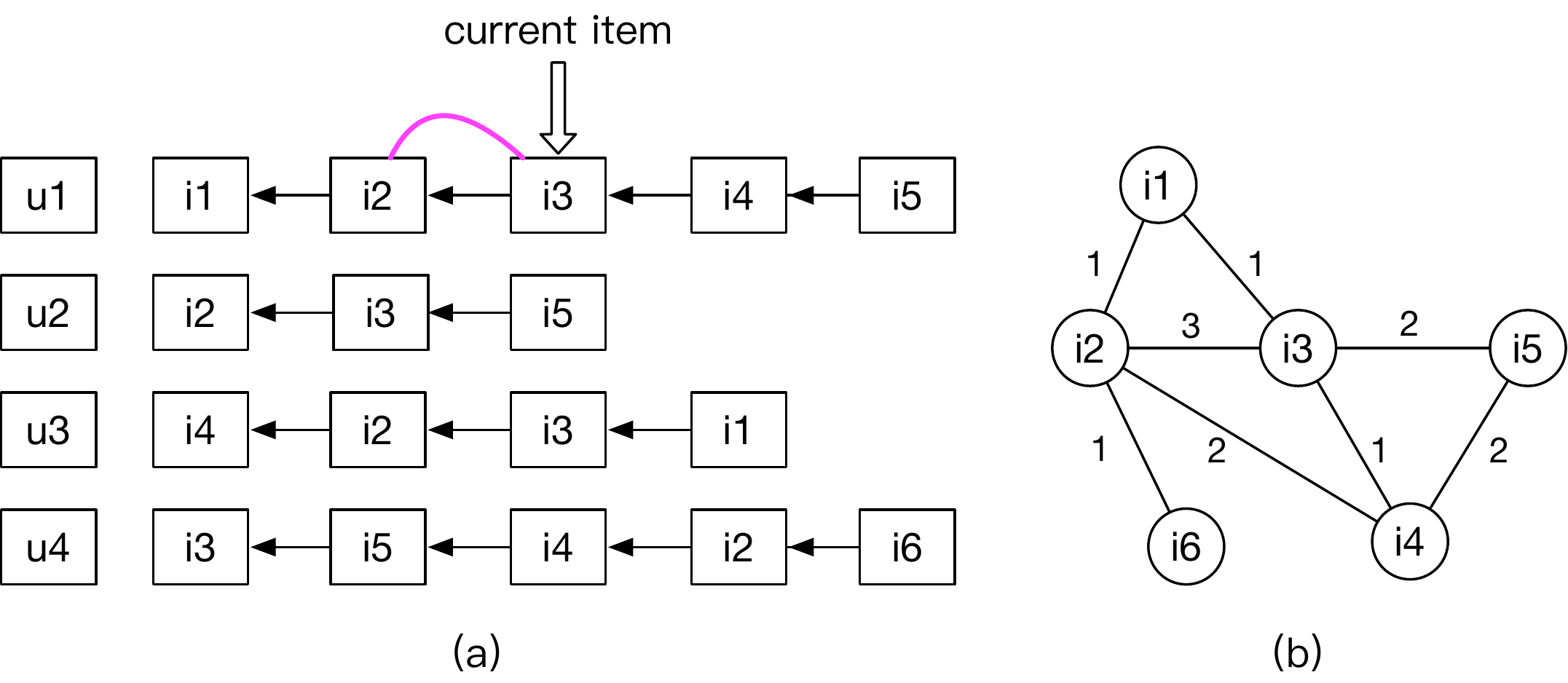}
\caption{The graph is constructed based on user history behaviors. (a) Each row represents a user's click sequence. The black arrow indicates the behavior direction, and the red arrow indicates the graph edge when the window size is 1. (b) In the co-occurrence commodity graph, nodes represent clicked commodities, and edge weights indicate the numbers of co-occurrence clicks.}
\label{fig:fig2}
\end{figure}

The detailed graph construction is shown in the Fig. \ref{fig:fig2}. Assuming the window size is 1, we construct an undirected edge to the left of each node in the sequence, and the co-occurrence commodity graph is obtained after processing each user's click sequence.

\subsection{Intention diffusion and aggregation}

\begin{figure}[!htbp]
\centering
\includegraphics[width=0.40\textwidth, keepaspectratio]{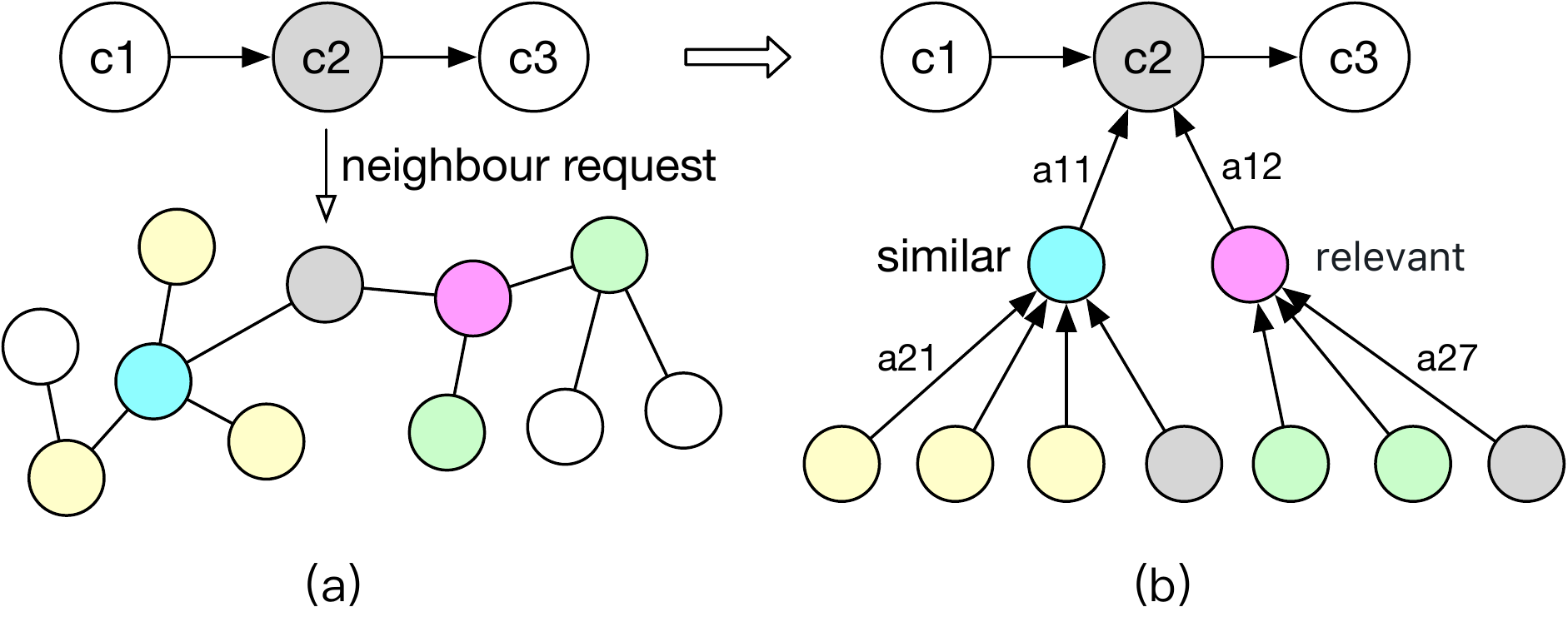}
\caption{The multi-layered intention diffusion and aggregation process is applied based on the co-occurrence commodity graph. Here c1, c2 and c3 represent user's click sequence. (a) indicates that the sequence of behavior is multi-layer diffusion into the graph. (b) indicates that the results of multi-layer diffusion are aggregated using the attention mechanism.}
\label{fig:fig3}
\end{figure}

We diffuse user's behavior sequence on co-occurrence graph to enrich user's intention expression as shown in Fig. \ref{fig:fig3}. Fig. \ref{fig:fig3}(a) contains user's behavior and co-occurrence commodity graph. Fig. \ref{fig:fig3}(b) is obtained by performing multi-layered neighbor diffusion on graph for each commodity of user click sequence. Then, The attention mechanism is applied to aggregate the tree-like intention.

\begin{figure}[!htbp]
\centering
\includegraphics[width=0.30\textwidth, keepaspectratio]{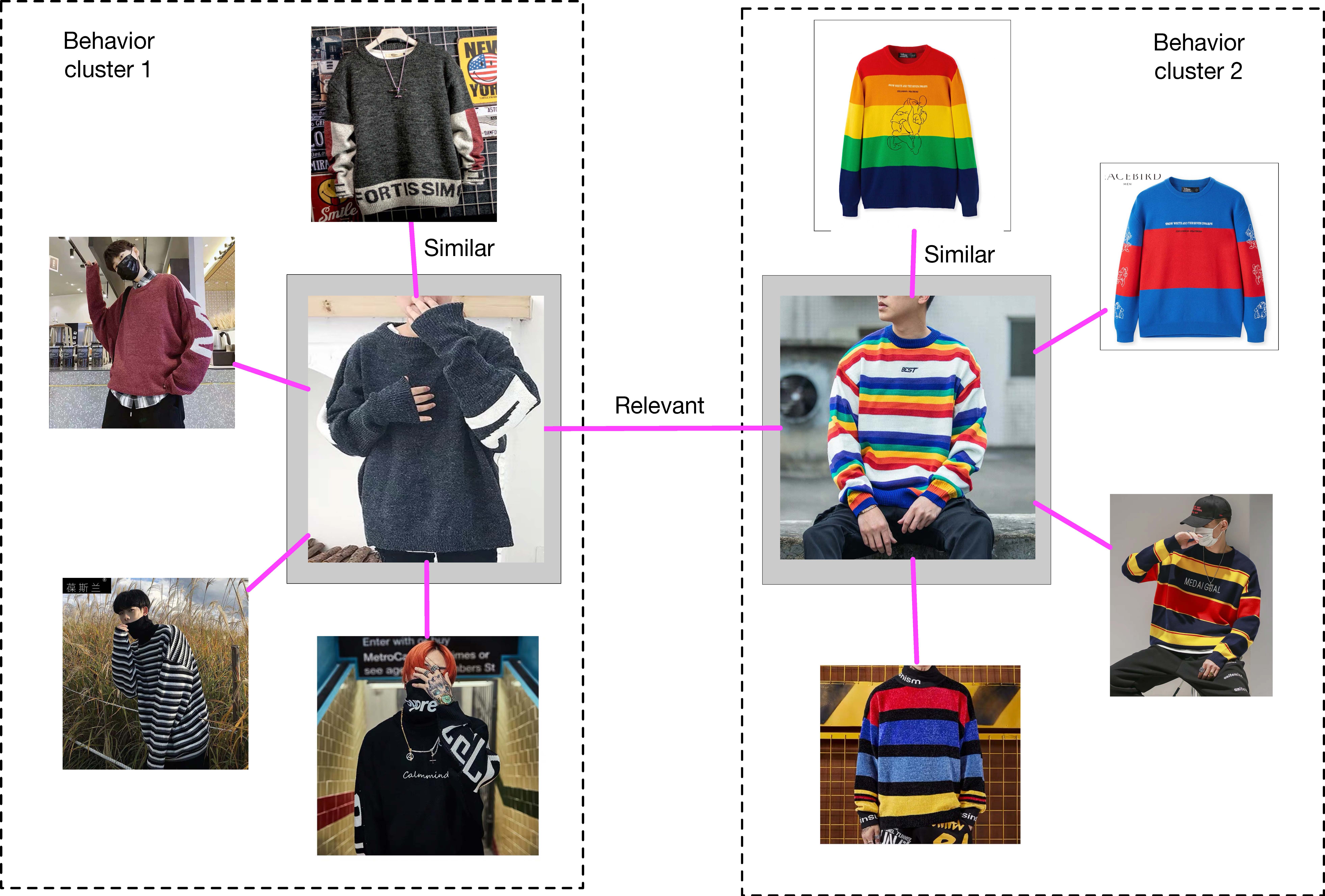}
\caption{The left and right parts represent two different behavior clusters with similar commodities in the co-occurrence commodity graph. The connection between cluster 1 and cluster 2 indicates strong relationship between two clusters, and the system utilizes these connections to help user jump out from cluster 1 to cluster 2 which is a potential preference.}
\label{fig:fig5}
\end{figure}

Diffusing user's real-time behavior on co-occurrence graph can recall two kinds of commodity. One is extremely similar commodities in same behavior cluster, it enriches user's behavior which is benefit for solving behavior sparsity problem. The other is relevant but not extremely similar commodities in another behavior cluster, which help user to jump out of their specific historical behavior for possible interest exploration, so the weak generalization problem is alleviated. Similar and relevant commodity is further described in Fig. \ref{fig:fig5}.

The intention diffusion and aggregation process is further detailed in Algorithm 1 inspired by \cite{velivckovic2017graph,ying2018graph}. We first diffuse each commodity of the user's click sequence in layers to explore commodities that have a strong co-occurrence relationship with the user's current click. Then aggregate the diffused commodities layer by layer from the outermost layer with AGGREGATE function described in Algorithm 2. Finally, we select the commodities of the user's click sequence by attention mechanism to increase the weight of the relevant commodities, and finally obtains the diverse vector including user’s potential preference.



\begin{algorithm}
\small
\caption{Graph Intention Discovery(GID)}
\begin{algorithmic}[1] 
  \Require
  Current  ranking ad $ad$,
  User click behaviors $pre\_clicks$,
  Depth $K$;
  Commodity similarity graph $G$;
  Neighbor select function $N$;
  \Ensure User implicit intention embedding $\bm{uii}$
  \State /* Implicit intention propagation */
  \State $S^{(K)} \gets pre\_clicks$
  \For{$k = K \to 1 $}
    \State $S^{(k-1)} \gets S^{(k)}$
    \For{$u \in S^{(k)}$}
      \State $S^{(k-1)} \gets S^{(k-1)} \bigcup N_G(u)$
    \EndFor
  \EndFor

  \State /* Implicit intention aggregation */
  \State $\bm{h_u^{(0)}} \gets \bm{x_u}, \forall u \in S^{(0)}$
  \For{$k=1 \to K$}
    \For{$ v \in S^{(k)}$}
      \State $ H \gets \{ \bm{h_u^{(k-1)}}, u \in N_G(v) \} $
      \State $\bm{h_v^{(k)}} \gets $AGGREGATE$^{(k)}(\bm{h_v^{(k-1)}}, H) $
    \EndFor
  \EndFor

  \State /* Generate user intention embedding by attention */
  \State $a_c \gets softmax(score(\bm{h_{ad}}, \bm{h_c^{(K)}})), \forall c \in S^{{K}} $
  \State $\bm{uii} \gets \sum_{c \in S^{(K)}} a_c \bm{h_c^{(K)}} $
\end{algorithmic}
\end{algorithm}
\vspace{-0.5cm}

\begin{algorithm}
\small
\caption{AGGREGATE}
\begin{algorithmic}[1] 
    \Require Current  node embedding $\bm{h_u}$ for node $u$,
             Set of neighbor embeddings $H=\{\bm{h_v}, v \in N_u\}$,
             Symmetric vector function $\gamma(\cdot)$
    \Ensure  AGGREGATED result $\bm{h_u^{aggr}}$ for node $u$
    \State  $  \alpha_{uv}=\frac{exp(ReLU(\bm{z^T}\cdot[\bm{Wh_u} || \bm{Wh_v}]))}{\sum_{k\in N_u}exp(ReLU(\bm{z^T}\cdot[\bm{Wh_u} || \bm{Wh_k}]))}, \forall v \in N_u\  $

    \State $\bm{n_u} \gets \gamma(\{$ReLU$(\bm{Mh_v}+\bm{m}) | v \in N_u\}, \bm{\alpha_u})$
    \State $\bm{h_u^{aggr}} \gets$ ReLU$(\bm{B} \cdot $CONCAT$(\bm{h_u}, \bm{n_u}) + \bm{b})$
\end{algorithmic}
\end{algorithm}
\vspace{-0.5cm}

\subsection{End-to-end joint training method}

The end-to-end joint training framework with graph-based intention mining and CTR prediction is shown in Fig. \ref{fig:fig1}. Firstly, we construct a co-occurrence commodity graph based on the method of Section 2.1, and use the graph engine euler\cite{renyi2018euler} to build a real-time graph neighbor query service. During the training phase, the multi-layered neighbor query is performed on the graph for each item in user's click sequence, and then the neighbors are aggregated according to the method described in 2.2 to obtain the intention vector. Secondly, this vector is concatenated with other features (e.g. query, user, ad and its statistical ctr) for CTR prediction.

In this process, the neighbor query, aggregation in graph and forward propagation are carried out in an end-to-end manner. Representation of graph node is updated by the back propagation algorithm based on the cross entropy loss defined in equation 1. The forward propagation process is further detailed in Algorithm 3.

\begin{algorithm}
\small
\caption{ Graph Intention Network }
\begin{algorithmic}[1] 
  \Require Set of samples with ($query$, $user$, $ad$, $pre\_clicks$);
  depth Parameter $K$;
  Forward propagation function $forward$;
  Commodity similarity graph $G$;
  Neighbor select function $N$;
  \Ensure Prediction of click-through rate $pctr$
  \State $ \bm{h} \gets $GID$(ad, pre\_clicks, K, G, N) $
  \State $ \textbf{features} \gets \text{CONCAT}(\{\bm{h_{\text{query}}}, \bm{h_{\text{user}}}, \bm{h_{\text{ad}}}, \bm{h}\}) $
  \State  $ \text{pctr} \gets \text{sigmoid}(\text{forward}(\textbf{features}))$
\end{algorithmic}
\end{algorithm}

\textbf{Loss function}:
The objective function of the joint training method is the cross entropy loss function as follows:
\begin{equation}
  L = -\frac{1}{N} \sum_{i=0}^{N}{ y_{i} \log(\text{pctr}_{i}) + (1 - y_{i}) \log(1 - \text{pctr}_{i})}
\end{equation}
Where $N$ is the total number of samples, $y_i$ is the label of the $i$th sample, and $\text{pctr}_{i}$ is the GIN forward propagation of the $i$th sample.

\section{Experiments}

To evaluate the performance of the proposed GIN method in CTR prediction tasks, we designed offline comparison experiments and further verified through online A/B testing.

\subsection{Experimental Setup}

\textbf{Graph data}:
The co-occurrence commodity graph is constructed using users' click behavior data during 30 days. There are 1 billion nodes and 8 billion edges. Types of graph node are all commodities. And the average output degree of graph node is 4.

\textbf{Train and Test data}:
Train data contains about 14 billion samples. Another 2 billion unseen samples are used to assess the performance of different CTR prediction models. Features include sparse id features and statistical features, corresponding to query, user, commodity, and historical behaviors.

\textbf{Competitors}:
We conduct experiments with several competitive methods on CTR modeling. (1) $\bm{Base}$: the baseline model for large scale CTR prediction task is neural factorization machines (NFM) which is widely used in industrial product. In this model, the sequence of user behavior is aggregated into an intention  vector by sum-pooling. (2) $\bm{DIN}$: This model uses the attention mechanism to weight the user behavior commodities and obtains the representation of user intention. (3) $\bm{GIN}$: The proposed method combines graph intention mining with CTR prediction task. The length of previous clicks is 20, and the depth parameter $K$ is set to 2. A 5-layer full-connection perceptron is adopted as the forward network with ReLU nonlinear activation. The neighbor is selected by the Top-$N$ function according to the edge weight.

\subsection{Offline evaluation}

The AUC is adopted as the offline performance metric. Higher AUC demonstrate better ranking performance. Same train and test data are used in these three CTR prediction models (Base, DIN, GIN).
The model effect is obtained after model parameters and optimizer configuration are all optimized.
Note that a 0.001 AUC increment means significantly performance improvement in our scenario.

The experimental results are shown in Table \ref{tab:tab2}.
Compared with DIN and Base, GIN has a significant effect increment.
DIN provides diverse intention expression through the attention mechanism, which improves the model ability to capture user's intention.
GIN further introduces the implicit intention information with graph diffusion, and solves the problems of behavior sparsity and weak generalization, which achieves the best CTR prediction performance.

\begin{table}
\small
  \caption{Comparison of effects of different models.}
  \label{tab:tab2}
  \begin{tabular}{cc}
    \toprule
    Method & Delta AUC \\
    \midrule
    DIN & +0.24\% \\
    GIN(ours) & \textbf{+0.60\%} \\
    \bottomrule
  \end{tabular}
  \vspace{-0.3cm}  
\end{table}

\begin{figure}[!htbp]
\centering
\includegraphics[width=0.48\textwidth, keepaspectratio]{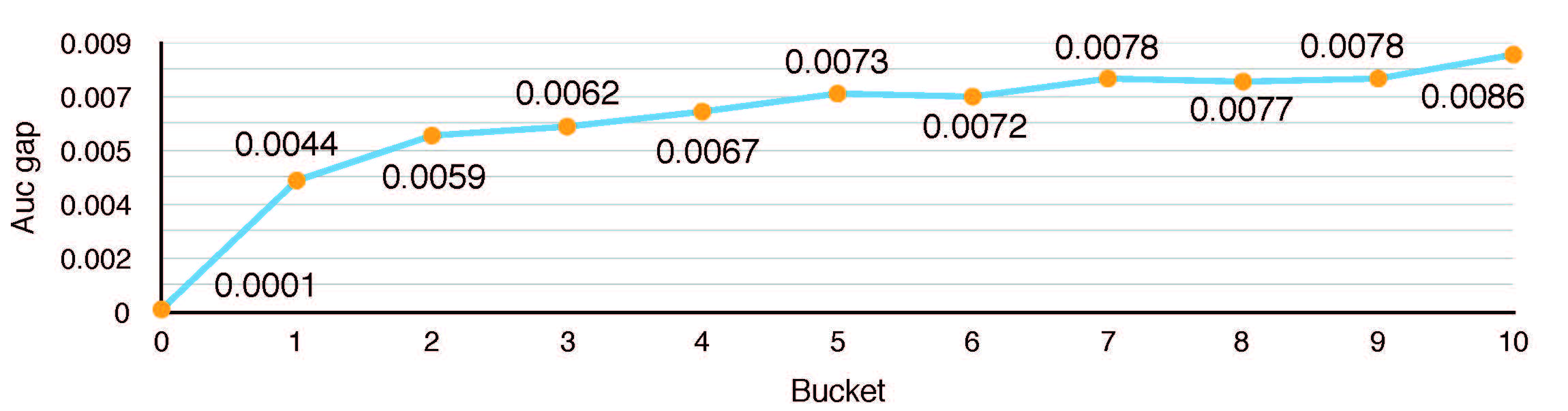}
\caption{Comparison of different behavior lengths.}
\label{fig:fig4}
\vspace{-0.3cm}
\end{figure}

Effect of GIN using different behavior lengths are shown in Fig. \ref{fig:fig4}.
The bucket id indicates different behavior lengths. AUC gap indicates GIN outperforms NFM significantly. The 0th bucket indicates GIN cannot perform the effect when there is no historical clicks.
For the case with less historical clicks, the effect of GIN has improved slightly, indicating that GIN has an effect on enriching user intention expression.
With more historical clicks, the improvement of GIN is more obvious. The reason may be that the user's intention is richly expressed. At the same time, it can discover user's potential preference to help user to migrate.

\begin{table}
\small
  \caption{Comparison of different neighbor numbers.}
  \label{tab:tab3}
  \begin{tabular}{ccc}
    \toprule
    \#Neighbors & Delta AUC & Time Cost \\
    \midrule
    GIN-5  & +0.39\% & 7h \\
    GIN-10 & +0.52\% & 12h \\
    GIN-20 & \textbf{+0.60\%} & 20h \\
    \bottomrule
  \end{tabular}
  \vspace{-0.3cm}  
\end{table}

\textbf{Neighbor number}:
In order to further explore the effect of the neighbor number on the model and the impact of each epoch time consumption, we compare the AUC changes and training time cost where the neighbor number is set as 3, 5, 10, 20 under K=2 conditions. As shown in Table \ref{tab:tab3}, as the number of neighbors increases, the effect becomes better, and the time cost also increases linearly. The reason is that the network overhead of the distributed system increases as the number of neighbors increases.

\textbf{Neighbor depth}:
We compare AUC and time cost for these CTR prediction models to explore the effect of intention diffusion within different neighbor depth as shown in Table \ref{tab:tab5}. GIN-0 means no neighbor info is utilized and GIN-2 means neighbors within two hop are aggregated. The AUC gap increases greatly as the depth grows, while time cost also increases rapidly.

\begin{table}
\small
  \caption{Comparison of different neighbor depth.}
  \label{tab:tab4}
  \begin{tabular}{ccc}
    \toprule
    Hop & Delta AUC & Time Cost \\
    \midrule
    GIN-order 1 & +0.45\% & \textbf{8h} \\
    GIN-order 2 & \textbf{+0.60\%} & 20h \\
    \bottomrule
  \end{tabular}
    \vspace{-0.3cm}  
\end{table}


\subsection{Online A/B Test}

We designed an online A/B test to further evaluate the performance of GIN. The comparisons of online CTR for different models during 3 consecutive days are shown in Table 4. The average CTR of GIN increased by 1.65\%,
indicating that GIN can effectively improve the effect of the CTR prediction task.

\begin{table}
\small
  \caption{Comparison of online CTR for 3 consecutive days.}
  \label{tab:tab5}
  \begin{tabular}{cccc}
    \toprule
    model & T & T+1 & T+2 \\
    \midrule
    DIN & +0.65\% & +0.66\% & +0.50\% \\
    GIN & \textbf{+1.46\%} & \textbf{+1.82\%} & \textbf{+1.67\%} \\
    \bottomrule
  \end{tabular}
      \vspace{-0.3cm}  
\end{table}

\section{Conclusion}

In this paper, we propose a novel approach GIN for CTR prediction in sponsored search. Using the end-to-end joint learning method of co-occurrence commodity graph and CTR prediction task, two important problems in user intention mining, i.e., behavior sparsity and weak generalization, are solved through the diffusion and aggregation of historical behaviors. Experiments on offline and online real-world dataset demonstrate the proposed GIN achieved excellent performance.

\bibliographystyle{ACM-Reference-Format}
\bibliography{kdd20}
\end{document}